\documentclass[runningheads]{llncs}
\usepackage[T1]{fontenc}
\usepackage{times}
\usepackage{soul}
\usepackage{url}
\usepackage[hidelinks]{hyperref}
\usepackage[utf8]{inputenc}
\usepackage[small]{caption} % 如果仍需调整标题大小，保留此项；否则删除
\usepackage{graphicx}
\usepackage{amsmath}
\usepackage{amssymb}
\usepackage{amsthm}
\usepackage{booktabs}
\usepackage{algorithm}
\usepackage{algorithmic}
\usepackage[switch]{lineno}
\usepackage{tabularx} 
\usepackage{multirow}
\usepackage{float}
\floatstyle{ruled}
\newfloat{algorithm}{tbp}{loa}[section]
\usepackage[caption=true,font=normalsize,labelfont=sf,textfont=sf]{subfig}
\usepackage{makecell} 
\usepackage{array}
\usepackage[backend=biber, style=numeric]{biblatex}
\begin{document}
\title{Whispers of Data: Unveiling Label Distributions in Federated Learning Through Virtual Client Simulation}
\titlerunning{Unveiling Label Distributions in Federated Learning}
\author{Zhixuan Ma\inst{1} \and
Haichang Gao$^*$\inst{1} \and
Junxiang Huang\inst{1} \and
Ping Wang \inst{1}}
\authorrunning{Ma et al.}
\institute{Xidian University, Xi'an Shannxi 710000, China 
\email{mazhixuan@stu.xidian.edu.cn}}
\maketitle              % typeset the header of the contribution
\begin{abstract}
Federated Learning enables collaborative training of a global model across multiple geographically dispersed clients without the need for data sharing. However, it is susceptible to inference attacks, particularly label inference attacks.

Existing studies on label distribution inference exhibits sensitive to the specific settings of the victim client and typically underperforms under defensive strategies. In this study, we propose a novel label distribution inference attack that is stable and adaptable to various scenarios. Specifically, we estimate the size of the victim client's dataset and construct several virtual clients tailored to the victim client. We then quantify the temporal generalization of each class label for the virtual clients and utilize the variation in temporal generalization to train an inference model that predicts the label distribution proportions of the victim client.

We validate our approach on multiple datasets, including MNIST, Fashion-MNIST, FER2013, and AG-News. The results demonstrate the superiority of our method compared to state-of-the-art techniques. Furthermore, our attack remains effective even under differential privacy defense mechanisms, underscoring its potential for real-world applications.

\keywords{Federated learning  \and Inference attack \and Data privacy \and Trust federated learning.}
\end{abstract}
\section{Introduction}

The traditional centralized deep learning approach requires aggregating distributed data in a central computing center for training. During data transmission, there is a risk that user data is stolen, leaked, or misused. Federated learning (FL) \cite{DBLP:conf/esorics/ChoHYLBP24,chen2024federated,huang2024federated,miao2024rfed}, as a distributed machine learning framework, enables geographically isolated and resource-constrained participants to securely collaborate on model training. In federated training, the participants' local data remain on their devices and are not shared with other participants or the central server. The model is trained by exchanging model parameters and performing global aggregation. Consequently, federated learning has become a widely adopted distributed machine learning technique, supporting privacy-sensitive domains such as healthcare \cite{nguyen2022federated,li2021federated}, finance \cite{long2020federated,imteaj2022leveraging}, and wireless communications \cite{RJOUB2025113574,AZHARSHOKOUFEH2025113526}.

A critical question in federated learning (FL) is whether private data is truly secure and trustworthy \cite{DBLP:conf/esorics/MozaffariCH24,DBLP:conf/esorics/AsareBKY23}. The model parameters uploaded by participants can inadvertently reveal characteristics of their local datasets \cite{10735243,DBLP:conf/esorics/GuepinMCM23,zhao2024loki}. Label distribution inference attacks, which expose a participant's preferences, represent a significant threat to the security of FL systems. For instance, a malicious healthcare institution could infer the prevalence of a particular disease, allowing it to stockpile medications and manipulate prices. Similarly, a malicious retailer could infer the distribution of certain products, thereby deducing supply and demand dynamics, potentially disrupting the market and gaining an unfair competitive advantage.

The label distribution inference attack \cite{ayora2021profiling,anelli2021put} originates from the Preference Profile Attack (PPA) \cite{PPA}, which was the first to propose inferring a participant's data preferences by observing the sensitivity of the model gradients uploaded by the participant. PPA identifies the top-k labels in which a target participant is most (or least) interested. Based on this framework, Raksha et al. \cite{ramakrishna2022inferring} and LDIA \cite{gu2023ldia} extend the approach to infer the entire label distribution by analyzing changes in the parameters of the model output layer of the target participant. However, the amplitude of these changes is influenced by factors such as the size of the victim client's dataset and the number of local training rounds, making it difficult for attackers to obtain prior knowledge of these parameters. Additionally, these methods are ineffective when differential privacy techniques are employed, as such techniques prevent an honest-but-curious server from accurately accessing the output layer parameters.

To eliminate sensitivity to changes in the model parameters, we leverage the temporal generalization of the model to train an inference model. This approach draws inspiration from the phenomenon of overfitting, where a model performs differently on training data compared to test data. This discrepancy occurs because the model tends to capture noise and details specific to the training data, thereby limiting its generalizability. Similarly, when the model is trained on labels with limited data, it often overfits by ``memorizing" the features of these samples rather than learning their underlying patterns. The scarcity of diverse samples for these labels hinders the model's ability to generalize, resulting in suboptimal performance on underrepresented labels. In contrast, labels with abundant data provide a richer variety of samples, allowing the model to better capture their general characteristics and improve its overall generalization performance. 

Based on the observation we proposed, we estimate the gradient information uploaded by the target client and design a virtual client environment that closely matches the target client's data size and distribution. By monitoring the temporal generalization performance of these virtual clients, we train an attack model to infer the local data distribution of the target client. Since inference defense strategies, such as differential privacy mechanisms, aim to balance model utility and privacy, our generalization performance-based inference attack is negligibly affected by these defense strategies. To achieve accurate label distribution inference, we introduce varying levels of noise into the virtual clients, effectively mitigating the impact of differential privacy mechanisms on the attack.

The main contributions of this paper are as follows.
\begin{itemize}
    \item We achieve a more accurate and stable label distribution inference attack by estimating the size of the dataset.

    \item Based on the estimated dataset size, we construct a virtual client cluster that simulates the target client's behavior in every probable distribution, including both IID and various non-IID scenarios.

    \item We quantify the generalization of each label then use the temporal generalization data as input and the label distribution proportion as output to train a time-series-based attack model.
    
    \item We evaluate the proposed attack model on four datasets, and demonstrating its effectiveness even under differential privacy defenses.
\end{itemize}

\begin{table*}[ht]
    \centering
    \caption{Comparison of three inference attacks.}
    \label{compare}
    \renewcommand\arraystretch{1.0} % Adjusts the height of each row for better visibility
    \begin{tabularx}{\textwidth}{|>{\centering\arraybackslash}p{2.5cm}|>{\centering\arraybackslash}p{2cm}|>{\centering\arraybackslash}p{3.4cm}|>{\centering\arraybackslash}X|} % Use p{} for fixed column width for the first three columns
        \hline
        \textbf{Category} & \textbf{Focus} & \textbf{Attack Target} & \textbf{Potential Impact} \\
        \hline
        \multirow{3}{*}{\makecell{Membership \\ Inference Attack}} & \multirow{3}{*}{Existence} & \multirow{3}{*}{\makecell{Determine the existence \\of  specific samples}} & Results in the disclosure of individual data users' private information. \\
        \hline
        \multirow{3}{*}{\makecell{Property \\ Inference Attack}} & \multirow{3}{*}{\makecell{Sensitive \\ Attributes}} & \multirow{3}{*}{\makecell{Extract specific sensitive \\ attributes of training data}} & Causes the leakage of critical attribute information, such as gender or medical conditions. \\
        \hline
        \multirow{7}{*}{\makecell{Label Distribution \\Inference Attack}} & \multirow{7}{*}{\makecell{Dataset \\Composition}} & \multirow{7}{*}{\makecell{Reveal the label \\distribution  across the\\ entire dataset}} & Attacking a single participant: Reveals individual or institutional preferences. \\
        & & & Attacking all participants: Exposes group traits, threatening collective privacy and fairness in decision. \\
        \hline
    \end{tabularx}
\end{table*}

\section{Related Works}

The label distribution inference attack aims to infer the label proportions in the training data of a target participant. By analyzing the model's outputs or statistics during the training process, the attacker seeks to determine the frequency or proportion of various class labels in the dataset. Zhou et al. first proposed the Preference Profiling Attack (PPA) \cite{PPA}, which infers the top $k$ categories of greatest or least interest to the target client by capturing gradient sensitivity. Subsequently, Raksha et al. \cite{ramakrishna2022inferring} and Gu et al. \cite{gu2023ldia} extended PPA from inferring a limited number of categories to inferring the label distribution across all categories. Their method simulated training by constructing auxiliary datasets and calculated changes in the output layer parameters of the trained model to deduce the label distribution ratios of the target participant's training data.
Dai et al. \cite{decaf} also analyzed the gradient of the last fully connected layer, leveraging its class-specific neuron activation patterns to identify null classes and decompose non-null class distributions via gradient bases constructed from auxiliary samples.
However, these approaches has notable limitations, as the inference results are heavily influenced by the auxiliary data set. Changes in model parameters following training on auxiliary datasets of varying sizes are fixed, and an honest-but-curious server cannot a priori know the target participant's dataset information, resulting in inaccurate predictions. Additionally, requiring participants to upload their complete local model parameters without any restrictions is impractical.
To enhance the security of participant models in federated learning, techniques such as differential privacy are often employed to protect uploaded information. However, these techniques also introduce noise, which contributes to the inaccuracy of existing methods in label proportion inference attacks.

Table \ref{compare} provides a comparative analysis of three inference attacks targeting federated learning. While all three methods exploit the data privacy of federated participants, they differ in their focus on data privacy, attack objectives, and potential impacts.
The \textbf{membership inference attack} aims to determine whether a specific data sample is included in the training set \cite{DBLP:conf/esorics/HoCSL24,DBLP:journals/tdsc/ZhengL24,DBLP:journals/tdsc/Pichler0VP24}. By analyzing the model's outputs, attackers can identify whether a user's sensitive data has been used in the training process. The success of this attack directly compromises individual privacy, exposing sensitive user information.
In contrast, the \textbf{attribute inference attack} seeks to infer specific characteristics or attributes of the samples, such as gender or age \cite{DBLP:conf/uss/AnnamalaiGR24,10662889}. Attackers analyze the model's prediction patterns in conjunction with auxiliary information to deduce sensitive attributes related to particular users. This type of attack not only undermines individual privacy but can also disproportionately affect certain societal groups, raising fairness concerns.
Finally, the \textbf{label distribution inference attack} focuses on understanding the distribution of various labels within the training dataset. By examining changes in model parameter sensitivity, attackers infer the frequency and proportions of labels present in the dataset. This attack may disclose sensitive information about participant preferences or group characteristics, potentially impacting the fairness and reliability of the federated model.

\section{Methodology}
\subsection{Threat Model}

\textbf{Victim Client:} This study examines the privacy risks associated with the label distribution of a target client. The client owns a private dataset $D$, which is stored locally and is not shared with any third party. In a federated learning system, the central server distributes the current global model $M_G$ to the client. The client then uses its local dataset $D$ to train $M_G$, producing an updated local model $M_L$. This local model, or the corresponding gradient update $G_L$, is subsequently uploaded to the server. Beyond this exchange, clients neither share their models with other participants nor provide interfaces for third-party access.

\textbf{Adversary Objective And Knowledge:} We assume the attacker is an honest-but-curious server. The goal of the honest-but-curious server is to infer the label distribution of participating clients' training data by analyzing the uploaded model parameters or update gradients during the federated learning process.
The server shares the same label space with clients but operates on different data spaces. The server has knowledge of the FL model's task and the class labels and can acquire auxiliary data corresponding to known labels from real-world sources. It  cannot manipulate the FL aggregation process, modify the aggregation strategy, or access the local data of any client directly or indirectly. The server can only observe the model parameters $M^t_L, t\in \{1,2,...T\}$ uploaded by the victim client during each training iteration.

\begin{figure*}[t]
\centering
    \includegraphics[width=\linewidth]{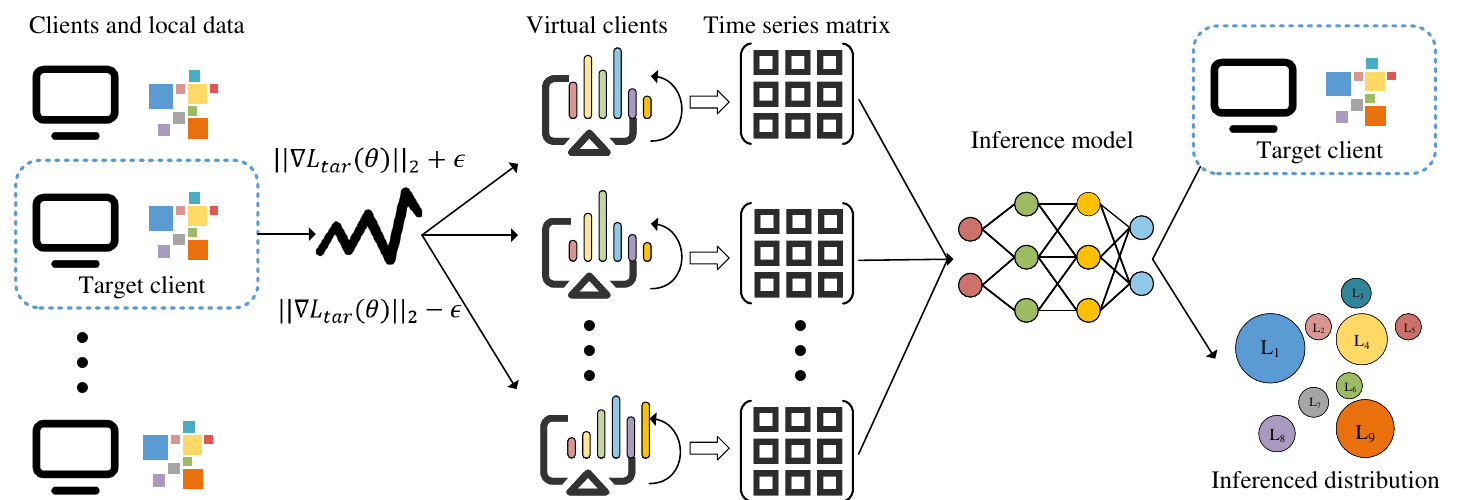}
    \caption{The overview of the framework.}
\label{framework}
\end{figure*}

\subsection{Overview}

The proposed method is summarized in Figure \ref{framework}. The honest-but-curious server aims to explore the data distribution privacy of participating clients without their awareness. The inference model is deployed on the server in two stages. The first stage estimates the size of the target participant's training dataset, and the second stage uses this estimation to construct virtual participants with a comparable dataset size. These virtual participants simulate federated training scenarios under various data distribution conditions. By collecting the temporal robustness accuracy changes during the training process of the virtual participants as input data and using their actual data distribution as labels, the server trains a inference model to probe the data distribution of the target participant.

\subsection{Estimation Size of Dataset}

In federated learning, clients are typically required to upload their dataset sizes to facilitate global convergence. However, to safeguard participant privacy, most federated learning strategies restrict clients from disclosing any details about their local datasets. Consequently, before inferring the label distribution of a target client, it is essential to first estimate the size of the client’s dataset.

\begin{figure}[ht]
    \centering
    \begin{minipage}[b]{0.4\linewidth} % Adjusted width of each minipage to fill the double column
        \includegraphics[width=\linewidth,height=0.87\linewidth]{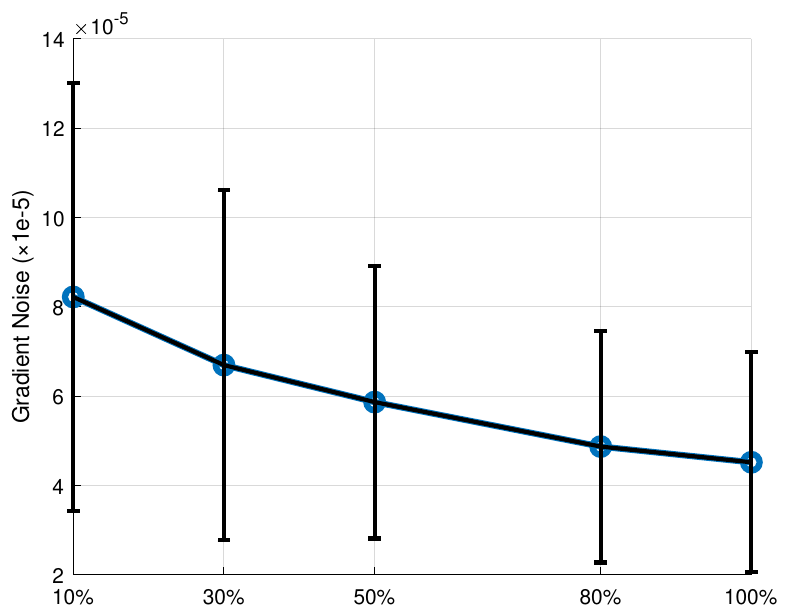}
        \caption*{gradient noise}
        \label{log}
    \end{minipage}
    \hfill
    \begin{minipage}[b]{0.4\linewidth}
        \includegraphics[width=\linewidth, height=0.87\linewidth]{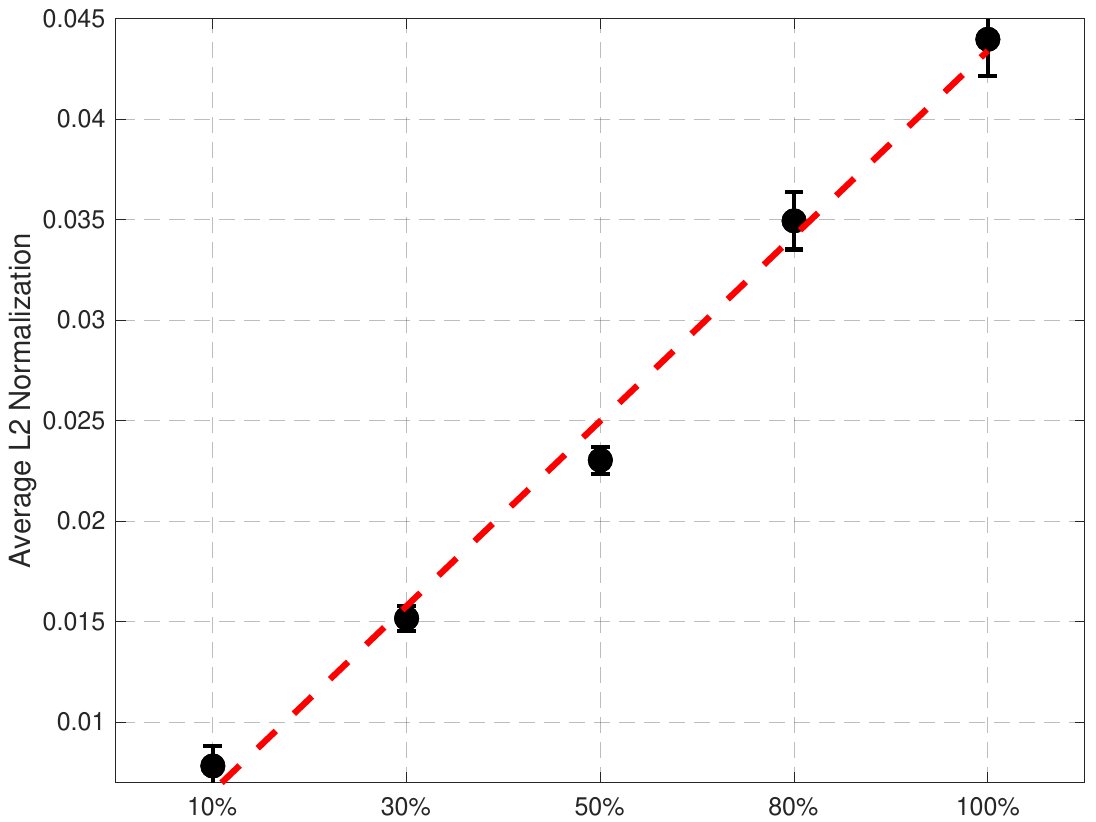}
        \caption*{gradient norm}
        \label{norm}
    \end{minipage}
    \caption{The influence of dataset size on model gradient.}
    \label{noise}
\end{figure}

We found that the size of the training dataset plays a critical role in determining the quality and stability of gradient updates in machine learning models. As shown in Fig \ref{noise}, we set the size of the client dataset to 2000(100\%), 1600(80\%), 1000(50\%), 600(30\%), and 200(10\%), recording the gradient noise and gradient norm during its training. Larger datasets generally lead to more stable and reliable gradient estimates. With a larger dataset, each batch better represents the overall data distribution, reducing variance in gradient estimates. 
In contrast, smaller datasets result in noisier gradient estimates due to increased variance, as each sample has a larger influence on the gradient. Although the magnitude of the gradient  may increase due to noisiness, the inconsistent directional alignment often leads to a smaller average $L_2$ norm ($||\nabla L(\theta)||_2$) resulting in oscillatory updates and slower convergence.

The relationship between dataset size and gradient variance can be quantified mathematically. For a gradient estimate $\hat{g}$ computed from a batch of size $B$, the variance is inversely proportional to $B$ and can be expressed as $Var(\hat{g}) = \frac{\sigma^2}{B}$, where $\sigma^2$ represents the variance of individual sample gradients. Smaller datasets, characterized by higher variance, produce noisier gradient estimates, increasing the difficulty of achieving stable updates. According to the Central Limit Theorem, when the batch size $B$ is sufficiently large, the mean gradient $\hat{g}$ computed from a batch approximates a normal distribution $\hat{g} \sim \mathcal{N}(\mu, \frac{\sigma^2}{B})$
where $\mu$ is the true gradient. For smaller datasets, the gradient estimates become noisier because of the higher variance introduced by fewer samples, which leads to more variability in the direction of the gradient. This results in unstable updates and difficulties in optimization. The noise in the gradient estimates can be described by $\epsilon$, where $\epsilon\sim \mathcal{N}(0,\frac{\sigma^2}n)$.
This additional noise amplifies variability in gradient direction, resulting in a less stable optimization trajectory.

The Taylor expansion of the loss function $L$ around a parameter $\theta$  further illustrates the impact of gradient stability on optimization: 
\begin{equation}
\begin{split}
    L(\theta+\Delta \theta) &\approx  L(\theta) + \nabla L(\theta) ^ \top\Delta \theta\\
    &\approx L(\theta)+\nabla L(\theta)^\top\Delta\theta+\frac{1}{2}\Delta \theta^\top H \theta
\end{split}    
\end{equation}
where $\nabla L(\theta)$ is the gradient, and $H$ is the Hessian matrix. The $L_2$ norm of the gradient quantifies the sensitivity of the loss function to changes in model parameters. For small datasets, high variance and inconsistent gradient directions reduce the $L_2$ norm:
\begin{equation}
\begin{split}
    ||\nabla L(\theta)||_2^2 = \sum_{i=1}^d (\nabla_i L(\theta))^2
\end{split}    
\end{equation}
where $\nabla_i L(\theta)$ is the $i$-th component of the gradient. When $\nabla L(\theta) = \mu + \epsilon$, the expected value of the squared $L_2$ norm is:
\begin{equation}
\begin{split}
    \mathbb{E}[||\nabla L(\theta)||_2^2] = ||\mu||_2^2 + \mathbb{E}[||\epsilon||_2^2]
\end{split}    
\end{equation}
The noise $\epsilon$ introduces additional variance, leading to oscillatory updates and slower convergence. Furthermore, the Hessian matrix $H$, which encodes the second-order curvature information of the loss function, f, depends on the dataset size. For dataset of size $n$, $H$ can be approximated as:
\begin{equation}
\begin{split}
    H \approx \frac{1}{n} \sum_{i=1}^n \nabla^2 L_i(\theta)
\end{split}    
\end{equation}
where $\nabla^2 L_i(\theta)$ represents the second-order partial derivatives for each sample. For smaller datasets, the estimate of the Hessian becomes less reliable, leading to a misrepresentation of the curvature and instability in optimization.

Finally, noisy gradients due to smaller datasets also lead to inconsistent directions during optimization. For a batch of size $B$, the average gradient is $\bar{g} = \frac{1}{B} \sum_{i=1}^B g_i$, and its $L_2$ norm is given by:
\begin{equation}
\begin{split}
   ||\bar{g}||_2 = \sqrt{\sum_{j=1}^d \left( \frac{1}{B} \sum_{i=1}^B g_{ij} \right)^2}
\end{split}    
\end{equation}
where $g_i$ represents the gradient for the $i$-th sample, and the sum is over the dimensions of the gradient vector. 
When individual gradients $g_i$ vary significantly in direction, cancellation effects reduce the aggregated gradient norm $||\bar{g}||_2$, leading to smaller effective step sizes and slower convergence.

In summary, the size of the training dataset critically influences gradient stability during optimization. Larger datasets reduce variance, stabilize gradient directions, and yield more consistent L2 norms, facilitating more efficient and reliable optimization. Conversely, smaller datasets introduce noisier gradients, higher variance, and directional inconsistencies, which hinder convergence and reduce the effectiveness of the training process.

Based on the above principles, we infer the dataset size of the target client by constructing virtual clients and comparing their $L_2$ norms with that of the target client. We collect real-world data with the same labels as the federated task to construct an auxiliary dataset $D_{aux}$. The virtual client is initialized with the same structure and parameters as the global model provided in the federated setting, ensuring consistency with the target client. During training, we record the gradient norm 
$||\nabla L_{v}(\theta)||_2$ of the virtual client. We employ a binary search method to iteratively adjust the size of the virtual client’s dataset until its gradient norm closely matches that of the target client, $||\nabla L_{tar}(\theta)||_2$. Specifically, we define a threshold $\epsilon$ to establish the acceptable range for the gradient norm, and the process continues until $||\nabla L_{v}(\theta)||_2$ satisfies the following condition:
\begin{equation}
    ||\nabla L_{tar}(\theta)||_2-\epsilon < ||\nabla L_{v}(\theta)||_2 < ||\nabla L_{tar}(\theta)||_2+\epsilon
\end{equation}
% The detailed process of such estimate is shown in Algorithm \ref{binary}.

\begin{algorithm}[!h]
    \caption{Estimating Dataset Size of the target Client.}
    \label{binary}

    \textbf{Input:} 
    Target client's gradient norm $||\nabla L_{\text{tar}}(\theta)||_2$, 
    Auxiliary dataset $D_{aux}$, 
    Global model parameters $\theta$, 
    Tolerance $\epsilon$.
    
    \textbf{Output:} 
    Estimated dataset size of the virtual client.
    
    \begin{algorithmic}[1]
        \STATE Initialize virtual client's model parameters $\theta_v$ with $\theta$
        \STATE Set an initial guess for dataset size $s_v$
        \STATE Define upper and lower bounds for the gradient norm:
               $\delta_1 = ||\nabla L_{\text{tar}}(\theta)||_2 - \epsilon$,
               $\delta_2 = ||\nabla L_{\text{tar}}(\theta)||_2 + \epsilon$
        \WHILE{$||\nabla L_v(\theta_v)||_2 < \delta_1$ or $||\nabla L_v(\theta_v)||_2 > \delta_2$}
            \STATE Train the virtual client with dataset of size $s_v$
            \STATE Compute the gradient norm $||\nabla L_v(\theta_v)||_2$
            \IF{$||\nabla L_v(\theta_v)||_2 < \delta_1$} 
                \STATE $s_v \leftarrow 2 \times s_v$
                \STATE Adjust the virtual client's dataset with $D_{aux}$
            \ELSIF{$||\nabla L_v(\theta_v)||_2 > \delta_2$}
                \STATE $s_v \leftarrow s_v / 2$
                \STATE Adjust the virtual client's dataset with $D_{aux}$
            \ENDIF
        \ENDWHILE
        \RETURN $s_v$
    \end{algorithmic}   
\end{algorithm}

\subsection{Construct The Virtual Clients and Inference Model}

To accurately simulate and train the inference model, we construct virtual clients under both IID and Non-IID scenarios. For the \textbf{IID scenario}, we ensure that each label is represented approximately equally across all virtual clients. The number of samples for each label is configured within the range $[C/s_v-\Delta,C/s_v+\Delta]$, where $C$ is the total number of classes, and $\Delta$ represents the allowable fluctuation range. 
For the \textbf{Non-IID scenario}, we consider two types of imbalances: label quantity-based imbalance and distribution-based label imbalance. In label quantity-based imbalance, each virtual client randomly selects samples from a fixed subset of $C_f$ classes. In distribution-based label imbalance, we employ a Dirichlet distribution to simulate uneven label distributions. The parameter $\alpha$ of the Dirichlet distribution controls the heterogeneity of the dataset. For each label distribution, we randomly sample from the auxiliary datasets to construct virtual client clusters $P$. By simulating federated training processes with these virtual clients and analyzing the changes in their generalization performance, we train the inference model to infer the label distribution of the target client.

\begin{algorithm}[!h]
    \caption{Construct Virtual Clients and Train Inference Model}
    \label{alg:inference_model}

    \textit{Initialize clusters of virtual clients $P$.}

    \textbf{Input:} 
    Auxiliary dataset $D_{aux}$, 
    Number of classes $C$, 
    Fluctuation range $\Delta$, 
    Dirichlet parameter $\alpha$, 
    Global model parameters $\theta$, 
    Number of training rounds $E$.

    \textbf{Output:} 
    Record $\mathbb{R}^{E \times C}$.

    \begin{algorithmic}[1]
        \FOR{each virtual client in $P$}
            \IF{IID scenario}
                \STATE $l \sim [C/s_v - \Delta, C/s_v + \Delta]$
            \ENDIF
            \IF{Non-IID scenario}
                \STATE $l \sim Dir(\alpha)$ or select $C_f$ classes
            \ENDIF
            \STATE Train the virtual clients $\hat{\theta}^{t+1} = \sum_{k=1}^N \frac{n_k}{n} \theta_k^{t}$
            \STATE Record $\mathbb{R}^{E \times C}$               

        \ENDFOR 
    \end{algorithmic}
    \textit{Inference the Label Distribution.}
    
    \textbf{Input:} $\mathbb{R}^{E \times C}$, $W_h$, $b_h$, $v$
    
    \textbf{Output:} $l=\{l_1, l_2, \dots, l_C\}$
    
    \begin{algorithmic}[1]
        \STATE $h_0, c_0 \leftarrow \text{Initialize}$
        \FOR{$t = 1$ to $E$}
            \STATE $h_t, c_t = \text{LSTM}(h_{t-1}, c_{t-1}, \mathbb{R}_{t,:})$
        \ENDFOR
        \STATE $e_t = v^\top \tanh(W_h h_t + b_h)$, $t = 1, \dots, E$
        \STATE $\alpha_t = \frac{\exp(e_t)}{\sum_{k=1}^E \exp(e_k)}$, $t = 1, \dots, E$
        \STATE $c = \sum_{t=1}^E \alpha_t h_t$
        \STATE $l = \text{FC}(c)$
        \RETURN $l$
    \end{algorithmic}
        % \nonumber \textit{Construct inference model.}
          
\end{algorithm}

The virtual client cluster $P$ operates independently of the federated learning process, meaning it does not participate in the sampling or aggregation phases. These virtual clients update their models directly based on the results from global aggregation. It is important to note that the virtual clients rely solely on global aggregation outcomes, ensuring that their operations do not interfere with the actual federated learning system. The attacker uses these virtual clients to record their performance on the test set at the end of each training round. For each virtual client, a temporal matrix $\mathbb{R}^{E \times C}$ is constructed, where $E$ represents the total number of training rounds. These temporal matrices capture the variations in the model’s generalization performance across different labels throughout the training process of the virtual clients.

To effectively extract key information and capture temporal dependencies from sequence data, we designed an LSTM-based inference model. The temporal matrices are used as inputs, while the output represents the label distribution proportions of the virtual clients, serving as the training targets for the inference model. Additionally, a temporal attention layer is incorporated before the output layer to enhance the model's ability to capture dependencies within the input sequences. Specifically, the intermediate output of the inference model is $H=[h_1,h_2,…, h_T]$, where $h_t$ denotes the hidden state at time step $t$. The temporal attention layer generates a weighted summary vector $c$ from these hidden states. To achieve this, we first compute an attention score $e_t$ to quantify the importance of each hidden state $e_t=v^\top tanh(W_hh_t+b_h)$, 
\begin{equation}
    e_t=v^\top tanh(W_hh_t+b_h)
\end{equation}
where $W_h$ and $b_h$ are learnable parameters, and $v$ is a parameter vector that projects the LSTM output into a scalar score. The attention scores are then normalized using the softmax function to generate weights $\alpha_t=\frac{exp(e_t)}{\sum^T_{i=1}exp(e_k)}$.
\begin{equation}
    \alpha_t=\frac{exp(e_t)}{\sum^T_{i=1}exp(e_k)}
\end{equation}
Finally, the context vector $c$ is obtained by computing a weighted average of the hidden states $h$ using the attention weights $c=\sum^T_{t=1}\alpha_th_t$. 
\begin{equation}
    c=\sum^T_{t=1}\alpha_th_t
\end{equation}
This LSTM model, enhanced with a temporal attention mechanism, is well-suited for handling the temporal performance matrices $\mathbb{R}^{E \times C}$. It effectively identifies and focuses on changes that impact the generalization performance of the virtual clients.

For the selected target client, the honest-but-curious server continuously collects its uploaded model parameters or gradients throughout the federated training process. Simultaneously, the server evaluates the target client's performance by obtaining its test accuracy using a public dataset. After multiple rounds of data collection, the temporal test data from the target client is fed into the trained inference model, yielding the predicted label distribution proportions.

\section{Experiments}
\label{4}

\subsection{Experimental Setup}
The proposed label distribution inference attack were conducted using PyTorch and Python 3 on an NVIDIA GeForce RTX 3090 equipped with 4 GPUs. We conducted comparisons of our proposed method with the SOTA under four different distance metrics. These comparisons were carried out in both the IID as well as non-IID scenarios.
For IID scenarios, we selected the MNIST \cite{MNIST}, Fashion-MNIST \cite{fashionMNIST}, and AG-News \cite{agnews} datasets for experimentation. For non-IID scenarios, in addition to the aforementioned datasets, we incorporated the Fer-2013 \cite{fer} dataset. We designed two non-IID situations to thoroughly evaluate the effectiveness of our method: one is label quantity-based imbalance, and the other is distribution-based label imbalance.

To emulate real-world data distributions, we constructed 1200 virtual clients under three different data distribution settings. The training group consisted of 900 virtual clients, while the testing group included 300 clients. Each virtual client was randomly assigned between 3,000 to 5,000 images to simulate varying levels of data availability. This setup allowed us to evaluate and compare the method's performance under different data distribution conditions in a controlled environment.

\textit{Evaluation metrics:} We assessed the proximity between the predicted and actual class-label distributions of target clients using four distance metrics: Wasserstein distance, KL divergence, JS divergence, and L1 distance, for a comprehensive analysis from multiple perspectives.

The \textbf{Wasserstein distance} assesses the structural and positional discrepancies between two distributions, calculating the minimal effort needed to reshape the predicted distribution to match the actual distribution.

\textbf{KL divergence} acts as an asymmetric metric for quantifying discrepancies between two probability distributions.

\textbf{JS divergence} provides a symmetric, bounded method for evaluating the overall similarity between two distributions, highlighting their global resemblance.

\textbf{L1 distance} quantifies the total absolute differences across corresponding dimensions of two distribution vectors, emphasizing the point-to-point disparities without considering their overall shapes or sample positions.

% \subsection{Inference Performance}

We evaluate the effectiveness of the proposed label distribution inference attack in IID and two non-IID scenarios. In the IID scenario, we set the data volume for each label to fluctuate within the range of $[0.9\times (N/C),1.1\times(N/C)]$, based on the total data volume $N$ and the number of classes $C$. This setup reflects the possible natural variation in data across clients in federated learning while avoiding a scenario where label distributions are completely uniform, as inferring label distribution proportions is meaningless when labels are perfectly balanced.
For the non-IID scenarios, we considered label quantity-based imbalance and distribution-based label imbalance. These conditions simulate common deviations in data distribution observed in the real world, where certain labels are overrepresented on some clients and relatively scarce on others. 

\subsection{Inference Performance in IID Scenario:} We conducted a comparative analysis of our technique against three SOTA approaches developed by Dai et al. (2024) \cite{decaf}, Raksha et al. (2022) \cite{ramakrishna2022inferring} and Zhou et al. (2023) \cite{PPA}. Specifically, in Zhou et al.'s method, the probabilities of predicting each class as the victim client's preferred label are normalized to estimate the proportions within the label distribution.

\begin{table}[ht]
    \centering
    \caption{Effectiveness of our model under different metrics in IID environment.}
    \label{table_iid}
    \renewcommand\arraystretch{1.0} % 行高压缩
    \scalebox{0.9}{ % 整体缩放系数
    \footnotesize % 字号缩小
    \begin{tabular}{|c|c|c|c|c|c|c|c|c|c|c|c|c|}
        \hline
        \multirow{2}{*}{\textbf{Method}} & 
        \multicolumn{4}{c|}{\textbf{MNIST}} & 
        \multicolumn{4}{c|}{\textbf{F-MNIST}} & 
        \multicolumn{4}{c|}{\textbf{AG-NEWS}} \\
        \cline{2-13}
        & \textbf{Wass} & \textbf{KL} & \textbf{JS} & \textbf{L1} & 
        \textbf{Wass} & \textbf{KL} & \textbf{JS} & \textbf{L1} & 
        \textbf{Wass} & \textbf{KL} & \textbf{JS} & \textbf{L1} \\
        \hline
        Dai et al. & 0.0441 & 0.0010 & 0.0003 & 0.0372 &0.0589 & 0.0015 & 0.0004 & 0.0444 & 0.0453 & 0.0016 & 0.0004 & 0.0493 \\
        \hline
        Raksha et al. & 0.2888 & 0.0733 & 0.0169 & 0.2871 & 0.3365 & 0.1257 & 0.0264 & 0.3441 & 0.1651 & 0.0223 & 0.0055 & 0.1771 \\
        \hline
        Zhou et al. & 2.9084 & 2.8086 & 0.3814 & 1.5061 & 2.7293 & 2.8857 & 0.3848 & 1.5267 & 1.0576 & 2.5025 & 0.2869 & 1.2998 \\
        \hline
        Ours & \textbf{0.0730} & \textbf{0.0015} & \textbf{0.0003} & \textbf{0.0438} & \textbf{0.0496} & \textbf{0.0017} & \textbf{0.0004} & \textbf{0.0440} & \textbf{0.0502} & \textbf{0.0030} & \textbf{0.0007} & \textbf{0.0626} \\
        \hline
    \end{tabular}}
\end{table}

% \begin{table}[ht]
%     \centering
%     \caption{Effectiveness of our model under different metrics in IID environment.}
%     \label{table_iid}
%     \renewcommand\arraystretch{1.25}
%     \scalebox{0.85}{
%     \begin{tabular}{|c|c|c|c|c|c|}
%         \hline
%         Dataset & Method & Wass & KL & JS & L1 \\
%         \hline
%         \multirow{3}{*}{MNIST} & Raksha et al. & 0.2888 & 0.0733 & 0.0169 & 0.2871 \\
%         & Zhou et al. & 2.9084 & 2.8086 & 0.3814 & 1.5061 \\
%         & Ours & \textbf{0.0730} & \textbf{0.0015} & \textbf{0.0003} & \textbf{0.0438} \\
%         \hline
%         \multirow{3}{*}{F-MNIST} & Raksha et al. & 0.3365 & 0.1257 & 0.0264 & 0.3441 \\
%         & Zhou et al. & 2.7293 & 2.8857 & 0.3848 & 1.5267 \\
%         & Ours & \textbf{0.0496} & \textbf{0.0017} & \textbf{0.0004} & \textbf{0.0440} \\
%         \hline
%         \multirow{3}{*}{AG-NEWS} & Raksha et al. & 0.1651 & 0.0223 & 0.0055 & 0.1771 \\
%         & Zhou et al. & 1.0576 & 2.5025 & 0.2869 & 1.2998 \\
%         & Ours & \textbf{0.0502} & \textbf{0.0030} & \textbf{0.0007} & \textbf{0.0626} \\
%         \hline
%     \end{tabular}}
% \end{table}

As illustrated in the Table \ref{table_iid}, the experimental results indicate that our model significantly outperforms three other advanced methods across all evaluated metrics.
Specifically, our approach exhibited exceptional performance on the MNIST and Fashion-MNIST datasets, where the low Wasserstein and L1 distances underscore the effectiveness of our inference model. On the AG-News dataset, despite the increased complexity of unstructured text data, our model maintained superior performance, particularly reflected in the low values of KL divergence and JS divergence, indicating that our model matches the actual distributions closely.

\begin{figure}[ht]
    \centering
    \begin{minipage}[b]{0.32\linewidth} % Adjusted width of each minipage to fill the double column
        \includegraphics[width=\linewidth,height=0.75\linewidth]{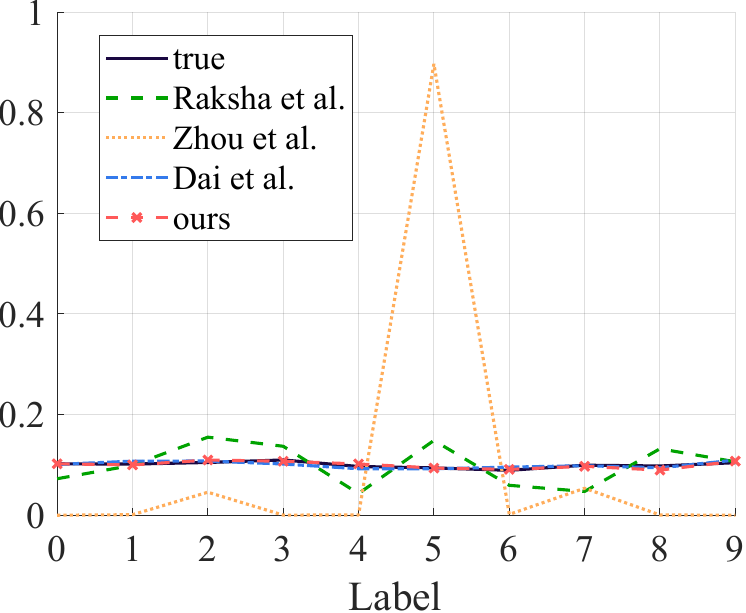}
        \caption*{MNIST}
        \label{1a}
    \end{minipage}
    \hfill
    \begin{minipage}[b]{0.32\linewidth}
        \includegraphics[width=\linewidth, height=0.75\linewidth]{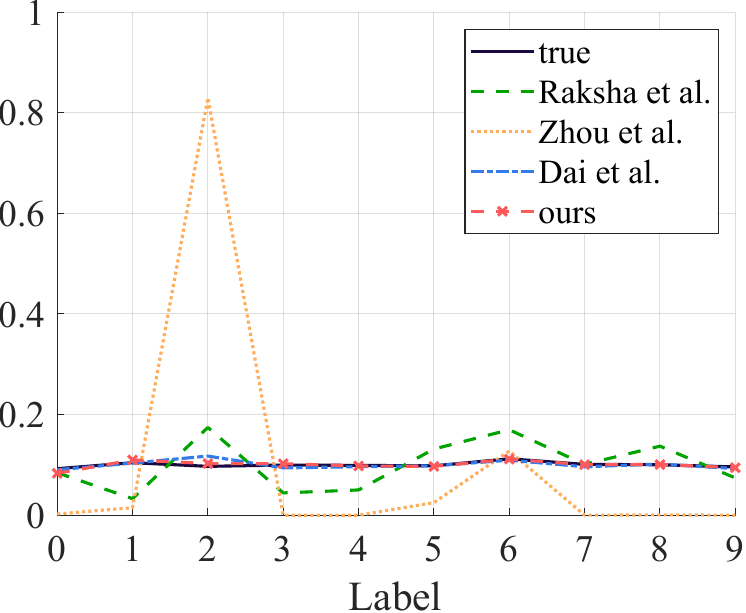}
        \caption*{Fashion-MNIST}
        \label{1b}
    \end{minipage}
    \hfill
    \begin{minipage}[b]{0.32\linewidth}
        \includegraphics[width=\linewidth, height=0.75\linewidth]{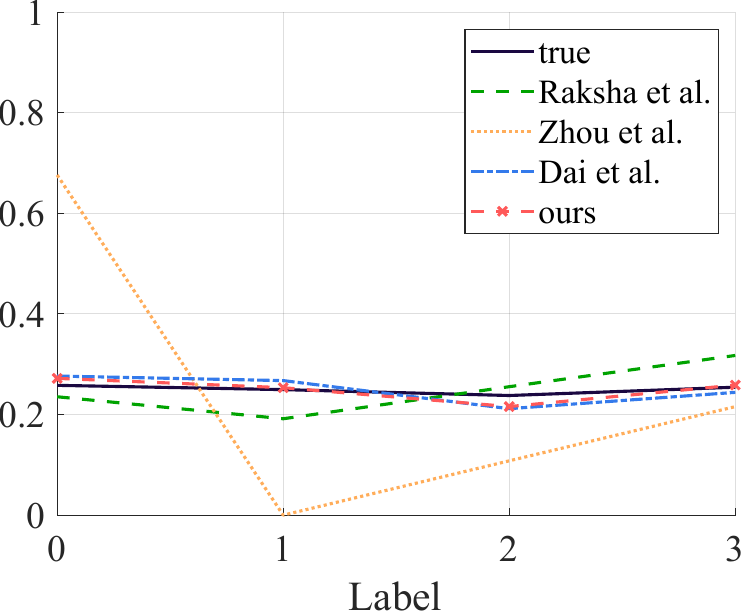}
        \caption*{AG-News}
        \label{1c} % Corrected label for unique referencing
    \end{minipage}
    \caption{Predicted distribution and true distribution in the case of IID.}
    \label{iid}
\end{figure}

We depicted the comparison between the predicted and actual label distributions as a line graph in Figure \ref{iid}. The analysis reveals that our method closely approximates the true label distributions across all datasets, with particularly notable accuracy on the MNIST and Fashion-MNIST datasets, where the predicted distributions almost perfectly align with the actual distributions. In contrast, although the method by Dai et al. and Raksha et al. performs well on certain labels, it exhibits considerable overall variability and inconsistency. The approach proposed by Zhou et al. often results in predictions that deviate significantly from the actual distributions.

\subsection{Inference Performance in Two Non-IID Scenario:} In federated learning, heterogeneity in data distribution presents a common and challenging issue. To verify the effectiveness of our proposed method within a federated learning environment, we conducted experiments in two non-IID scenarios: Label Quantity-Based Imbalance and Distribution-Based Label Imbalance. 

\textbf{Label quantity-based imbalance} refers to significant disparities in the number of samples for each class label across different clients. In this scenario, each client possesses a varying number of samples for each class, while the total number of class labels $C$ remains fixed. For datasets with a larger number of labels, such as MNIST, Fashion-MNIST, and Fer2013 we set C = 3. For the AG-News dataset, we set C = 2.

\begin{figure*}[ht]
    \centering
    \begin{minipage}[b]{0.24\linewidth} % Adjusted width of each minipage to fill the double column
        \includegraphics[width=\linewidth,height=0.87\linewidth]{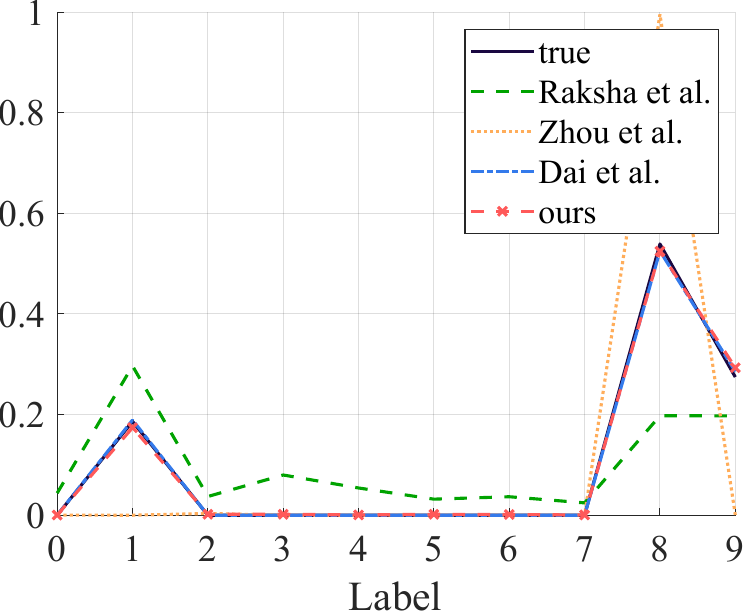}
        \caption*{MNIST}
        \label{1a}
    \end{minipage}
    \hfill
    \begin{minipage}[b]{0.24\linewidth}
        \includegraphics[width=\linewidth, height=0.87\linewidth]{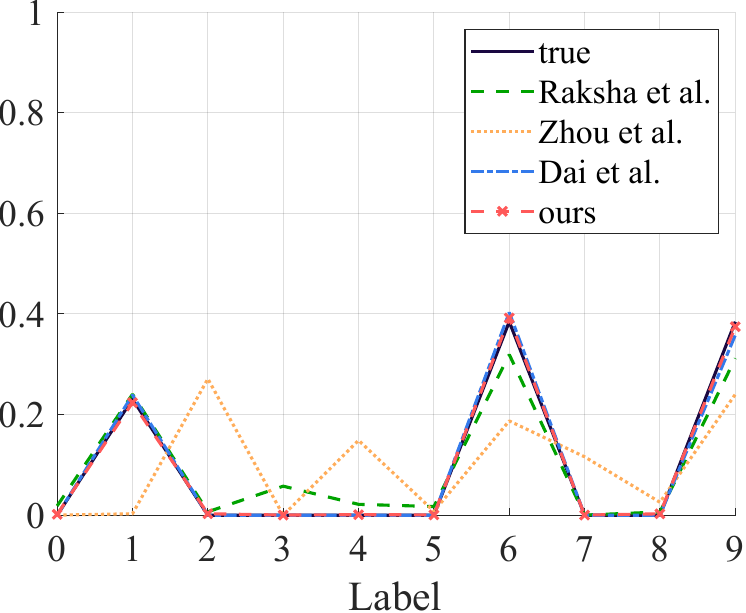}
        \caption*{Fashion-MNIST}
        \label{1b}
    \end{minipage}
    \hfill
    \begin{minipage}[b]{0.24\linewidth}
        \includegraphics[width=\linewidth, height=0.87\linewidth]{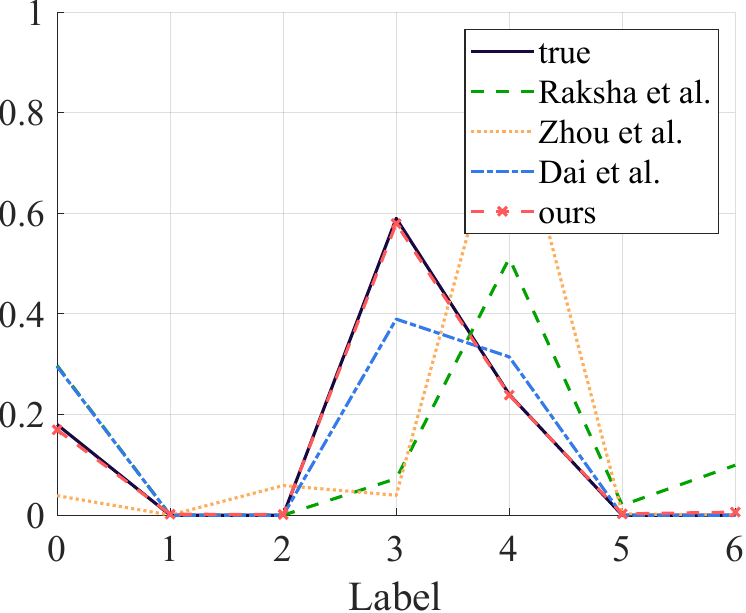}
        \caption*{Fer2013}
        \label{1c}
    \end{minipage}
    \hfill
    \begin{minipage}[b]{0.24\linewidth}
        \includegraphics[width=\linewidth, height=0.87\linewidth]{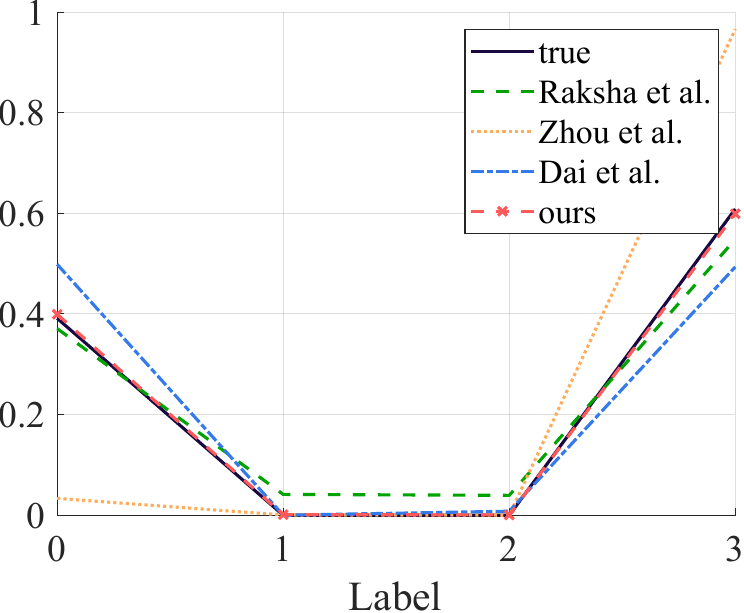}
         \caption*{AG-News}
        \label{1d} % Corrected label for unique referencing
    \end{minipage}
    \caption{Predicted distribution and true distribution in the case of Non-IID of label quantity-based imbalance.}
    \label{fig_c}
\end{figure*}

The real label distribution proportions of the victim client compared to the predicted distributions are illustrated in the Figure \ref{fig_c}. Our method exhibits significant advantages over the approaches by Dai et al., Raksha et al. and Zhou et al. across four datasets. Our model more accurately approximates the true distributions, particularly in scenarios where label distributions are highly concentrated or highly dispersed. On the MNIST and Fer2013 datasets, our model precisely captures the distribution of high-frequency labels, with the predicted and actual label distribution graphs nearly overlapping. 
Although the methods by Dai et al. and Raksha et al. can approximate the true distribution in some datasets, they tend to either overestimate or severely underestimate the proportions of certain labels in most cases. Overall, our model demonstrates extensive adaptability and robustness in complex label imbalances scenarios. This capability underscores its potential in practical federated learning applications, providing strong support and accurate analytical foundations for real-world applications with uneven label quality.

\textbf{Distribution-based label imbalance} concerns the feature distribution within each label, which may vary drastically across different clients. To simulate this data distribution, we employ a Dirichlet distribution to partition data across clients and adjust the heterogeneity level among clients by manipulating the Dirichlet parameter $\alpha$. The  smaller $\alpha$ values indicating high client data heterogeneity, while larger $\alpha$ values lead to a more uniform distribution. We set $\alpha =\{ 0.5, 1, 2 \} $ for MNIST, F-MNIST, and FER2013, $\alpha =\{0.1, 0.4, 1\}$ for AG-NEWS.

\begin{table}[h]
\centering
\caption{Effectiveness of our model under different metrics in Non-IID environment of distribution-based label imbalance.}
\label{table_d}
\scalebox{0.675}{
\renewcommand{\arraystretch}{1.5} % 增加行高
\begin{tabular}{|*{18}{c|}} % 18列，每列居中对齐
\hline
\multirow{2}{*}{Method} & \multirow{2}{*}{$\alpha$} & \multicolumn{4}{c|}{MNIST} & \multicolumn{4}{c|}{F-MNIST} & \multicolumn{4}{c|}{FER2013} & \multicolumn{4}{c|}{AG-NEWS(0.1/\ 0.4/\ 1)} \\
\cline{3-18}
 &  & Wass &KL &JS &L1 & Wass &KL &JS &L1 & Wass &KL &JS &L1 & Wass &KL &JS &L1 \\ % 空行
\hline
Dai et al. & \multirow{4}{*}{\makecell{0.5\\ /\ \\0.1}} & 0.1508 & 0.0282 & 0.0052 & 0.0919 & 0.1008 & 0.0252 & 0.0052 & 0.0849 & 0.4800 & 0.1391 & 0.0363 & 0.4141 & 0.1183 & 0.0507 & 0.0148 & 0.1522 \\
\cline{1-1}\cline{3-18}
Raksha et al. &  & 0.9071 & 0.2609 & 0.0694 & 0.5338 & 0.4425 & 0.0847 & 0.0220 & 0.3121 & 0.6742 & 0.2632 & 0.0665 & 0.5502 & 0.4510 & 0.2846 & 0.0864 & 0.5283 \\
\cline{1-1}\cline{3-18}
Zhou et al. &  & 1.7926 & 2.0411 & 0.2451 & 1.1215 & 2.1623 & 2.4024 & 0.3052 & 1.3042 & 1.2926 & 1.7846 & 0.2240 & 1.0706 & 0.1434 & 0.1962 & 0.0295 & 0.2163 \\
\cline{1-1}\cline{3-18}
Ours &  & \textbf{0.1211} & \textbf{0.0149} & \textbf{0.0041} & \textbf{0.0811} & \textbf{0.0837} & \textbf{0.0085} & \textbf{0.0024} & \textbf{0.0634} & \textbf{0.0972} & \textbf{0.0283} & \textbf{0.0073} & \textbf{0.0972} & \textbf{0.0524} & \textbf{0.0163} & \textbf{0.0049} & \textbf{0.0602} \\
\hline
Dai et al. & \multirow{4}{*}{\makecell{1 \\ /\ \\0.4}} & 0.1230 & 0.0352 & 0.0059 & 0.0913 & 0.1098 & 0.0192 & 0.0042 & 0.0853 & 0.2972 & 0.1005 & 0.0213 & 0.2855 & 0.2898 & 0.1199 & 0.0323 & 0.3286 \\
\cline{1-1}\cline{3-18}
Raksha et al. &  & 0.4884 & 0.0846 & 0.0226 & 0.3159 & 0.2962 & 0.0586 & 0.0141 & 0.2373 & 0.4807 & 0.2937 & 0.0490 & 0.4400 & 0.5214 & 0.2833 & 0.0780 & 0.5815 \\
\cline{1-1}\cline{3-18}
Zhou et al. &  & 1.7856 & 1.8181 & 0.2362 & 1.1302 & 2.3492 & 2.2553 & 0.2976 & 1.3033 & 1.6383 & 2.2295 & 0.2875 & 1.2653 & 0.5403 & 0.8894 & 0.1080 & 0.5978 \\
\cline{1-1}\cline{3-18}
Ours &  & \textbf{0.0937} & \textbf{0.0067} & \textbf{0.0018} & \textbf{0.0680} & \textbf{0.0848} & \textbf{0.0080} & \textbf{0.0022} & \textbf{0.0694} & \textbf{0.0689} & \textbf{0.0081} & \textbf{0.0020} & \textbf{0.0799} & \textbf{0.0909} & \textbf{0.0166} & \textbf{0.0043} & \textbf{0.1082} \\
\hline
Dai et al. & \multirow{4}{*}{\makecell{2\\ /\ \\1}} & 0.0041 & 0.0275 & 0.0050 & 0.1026 & 0.1152 & 0.0255 & 0.0050 & 0.0934 & 0.2476 & 0.0878 & 0.0178 & 0.2503 & 0.2823 & 0.1039 & 0.0286 & 0.3494 \\
\cline{1-1}\cline{3-18}
Raksha et al. &  & 0.3527 & 0.0504 & 0.0130 & 0.2204 & 0.2803 & 0.0827 & 0.0140 & 0.2073 & 0.6231 & 0.3442 & 0.0628 & 0.5140 & 0.3667 & 0.1669 & 0.0462 & 0.4348 \\
\cline{1-1}\cline{3-18}
Zhou et al. &  & 2.3480 & 2.1417 & 0.2836 & 1.2770 & 2.2807 & 2.2660 & 0.3050 & 1.3314 & 1.6760 & 2.2675 & 0.2828 & 1.2595 & 1.2595 & 1.7653 & 0.2176 & 1.0345 \\
\cline{1-1}\cline{3-18}
Ours &   & \textbf{0.0929} & \textbf{0.0075} & \textbf{0.0021} & \textbf{0.0706} & \textbf{0.0896} & \textbf{0.0047} & \textbf{0.0012} & \textbf{0.0613} & \textbf{0.0892} & \textbf{0.0072} & \textbf{0.0018} & \textbf{0.0883} & \textbf{0.0067} & \textbf{0.0124} & \textbf{0.0027} & \textbf{0.1067} \\
\hline
\end{tabular}}
\end{table}

The Table \ref{table_d} illustrates the effectiveness of our method compared to SOTA under a non-IID setting with distribution-based label imbalance. Across all four measurement metrics, our method consistently shows lower distances on all datasets, underscoring the close alignment between the predicted and actual label distributions. This indicates that our approach effectively captures both the general and specific features of the label distribution. Notably, our method reduces the L1 distance by an order of magnitude compared to alternative methods. On the MNIST dataset with an alpha value of $2$, the KL divergence between our inferred label distribution and the true distribution is just $0.0027$, maintaining a high degree of similarity between the predicted and actual distributions.

\begin{figure*}[h]
    \centering
    % First row of images
    \begin{minipage}[b]{0.24\linewidth}
        \includegraphics[width=\linewidth]{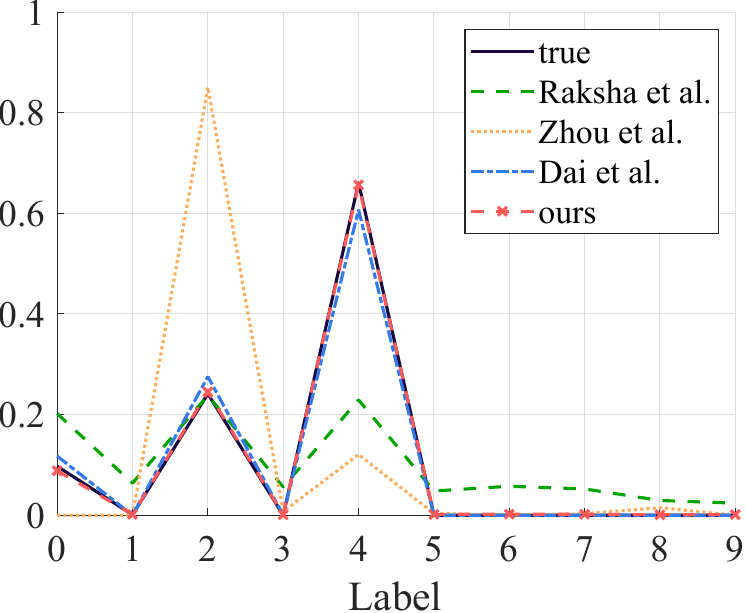}
        \caption*{$\alpha=0.5$}
        \label{mnist05}
    \end{minipage}
    \hfill
    \begin{minipage}[b]{0.24\linewidth}
        \includegraphics[width=\linewidth]{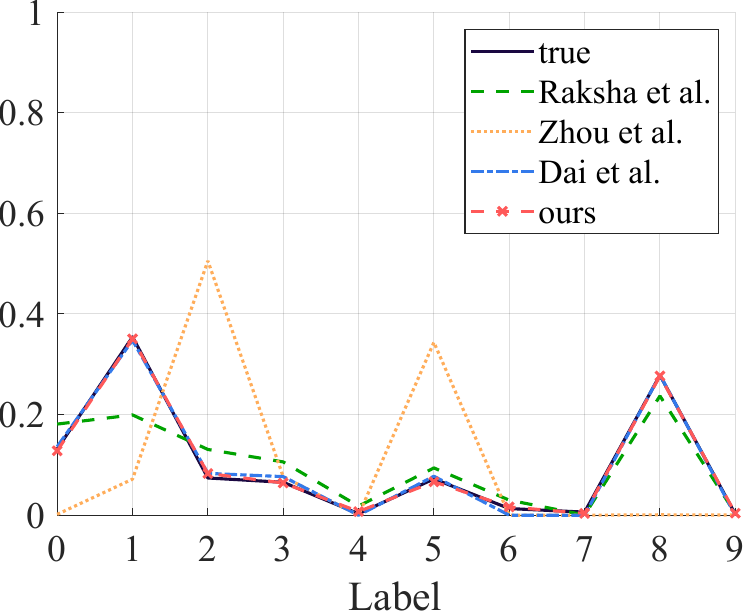}
        \caption*{$\alpha=0.5$}
        \label{fmnist05}
    \end{minipage}
    \hfill
    \begin{minipage}[b]{0.24\linewidth}
        \includegraphics[width=\linewidth]{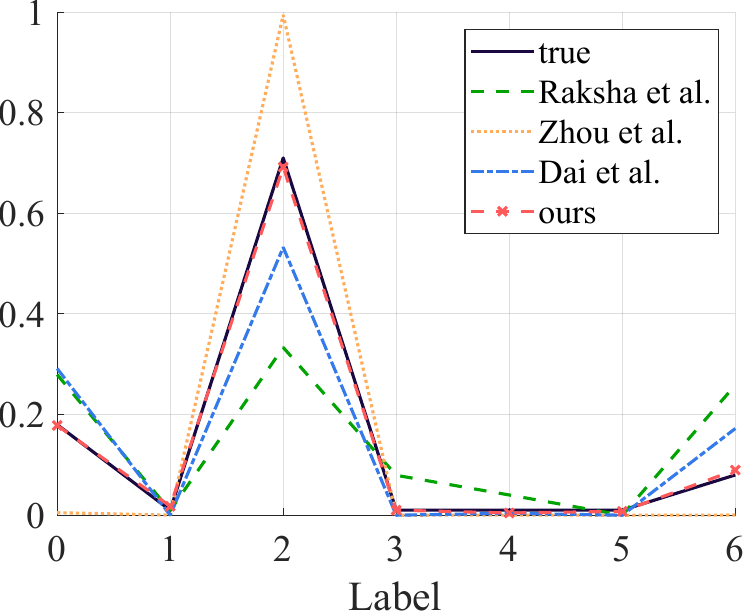}
        \caption*{$\alpha=0.5$}
        \label{fer05}
    \end{minipage}
    \hfill
    \begin{minipage}[b]{0.24\linewidth}
        \includegraphics[width=\linewidth]{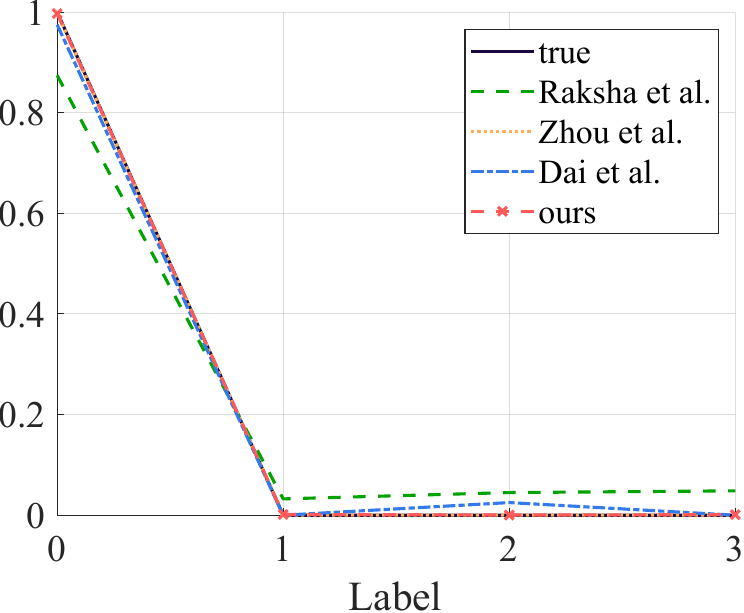}
        \caption*{$\alpha=0.1$}
        \label{agnews01}
    \end{minipage}

    \vspace{1em} % Adds some space between the rows

    % Second row of images
    \begin{minipage}[b]{0.24\linewidth}
        \includegraphics[width=\linewidth]{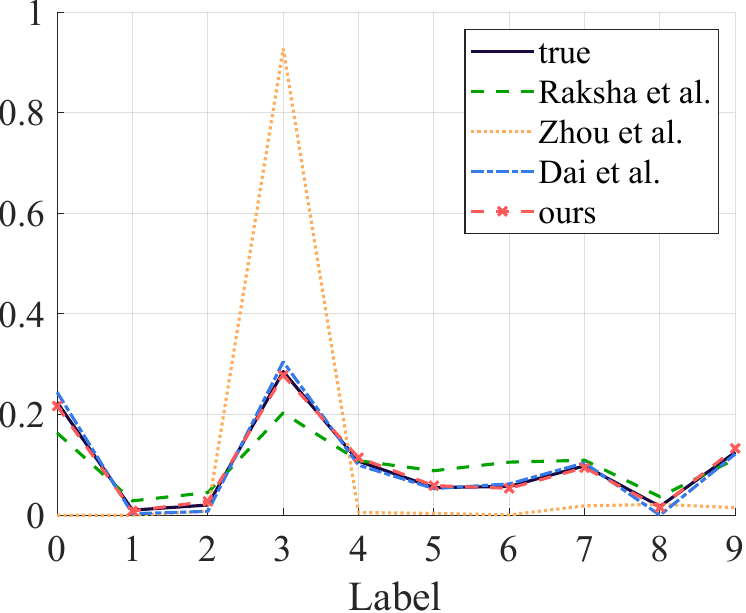}
        \caption*{$\alpha=1$}
        \label{mnist1}
    \end{minipage}
    \hfill
    \begin{minipage}[b]{0.24\linewidth}
        \includegraphics[width=\linewidth]{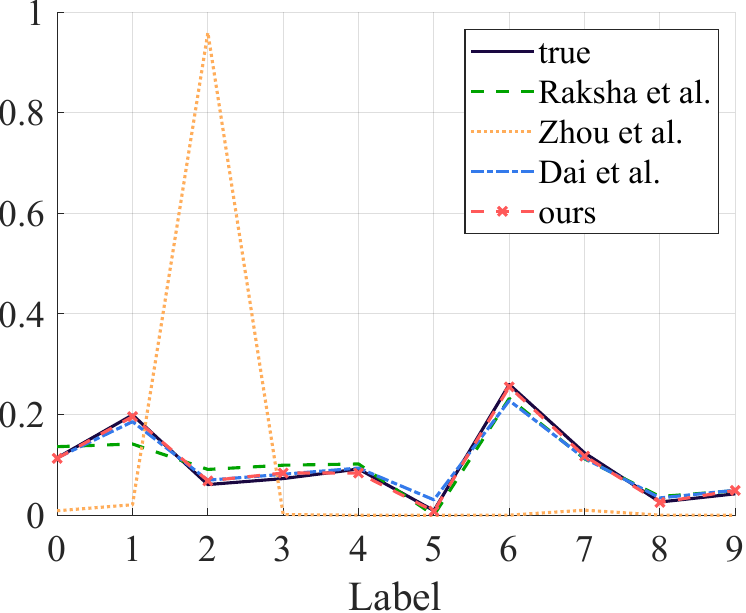}
        \caption*{$\alpha=1$}
        \label{fmnist1}
    \end{minipage}
    \hfill
    \begin{minipage}[b]{0.24\linewidth}
        \includegraphics[width=\linewidth]{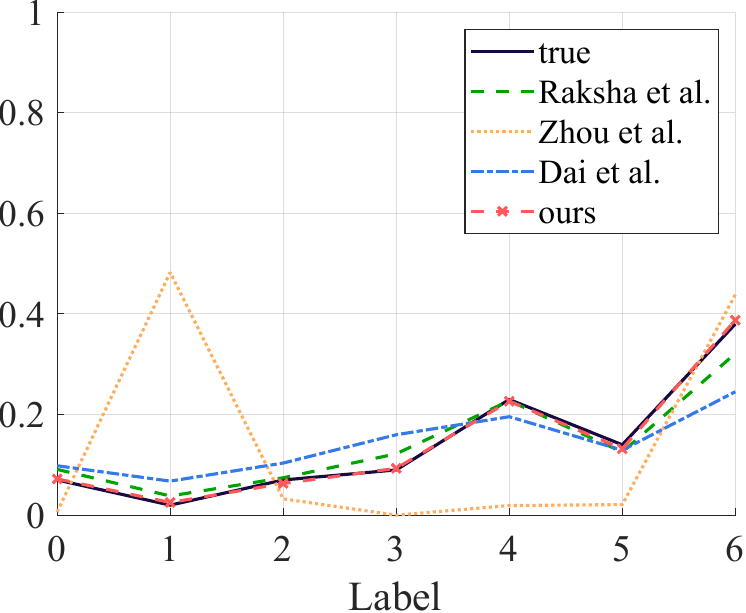}
        \caption*{$\alpha=1$}
        \label{fer1}
    \end{minipage}
    \hfill
    \begin{minipage}[b]{0.24\linewidth}
        \includegraphics[width=\linewidth]{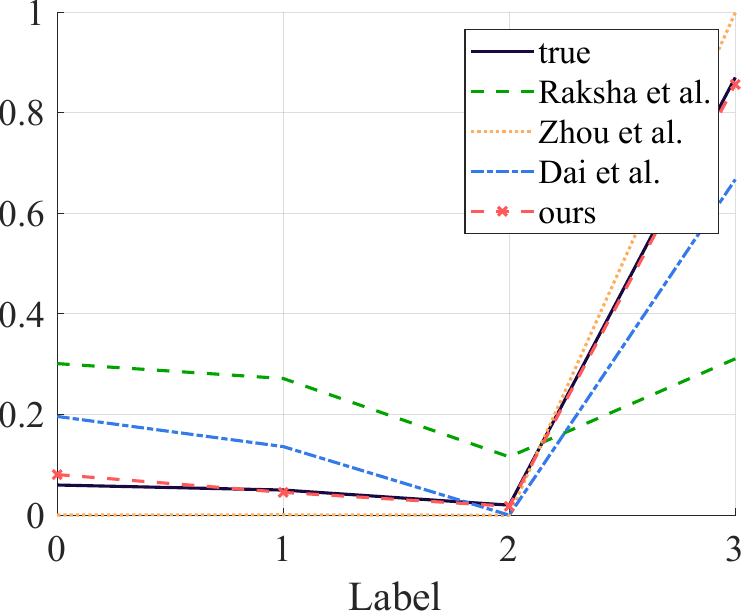}
        \caption*{$\alpha=0.4$}
        \label{agnews04}
    \end{minipage}

    \vspace{1em} % Adds some space between the rows

    % Third row of images
    \begin{minipage}[b]{0.24\linewidth}
        \includegraphics[width=\linewidth]{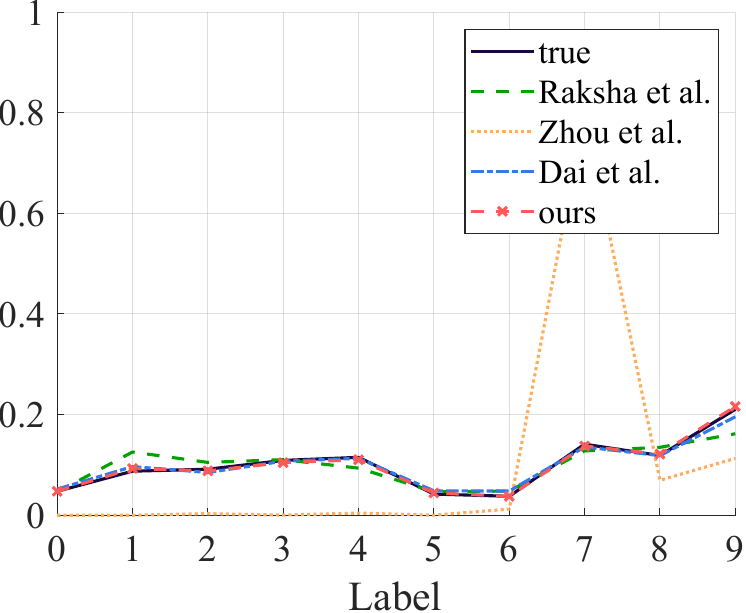}
        \caption*{$\alpha=2$}
        \label{mnist2}
    \end{minipage}
    \hfill
    \begin{minipage}[b]{0.24\linewidth}
        \includegraphics[width=\linewidth]{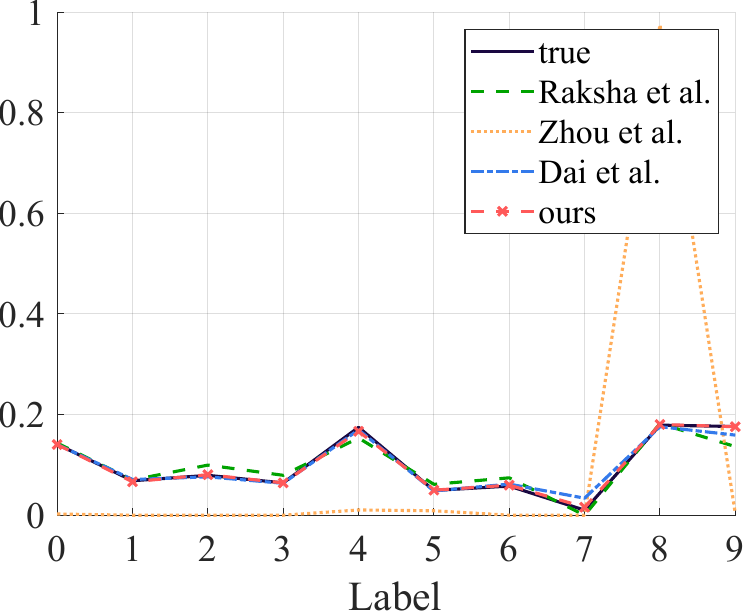}
        \caption*{$\alpha=2$}
        \label{fmnist2}
    \end{minipage}
    \hfill
    \begin{minipage}[b]{0.24\linewidth}
        \includegraphics[width=\linewidth]{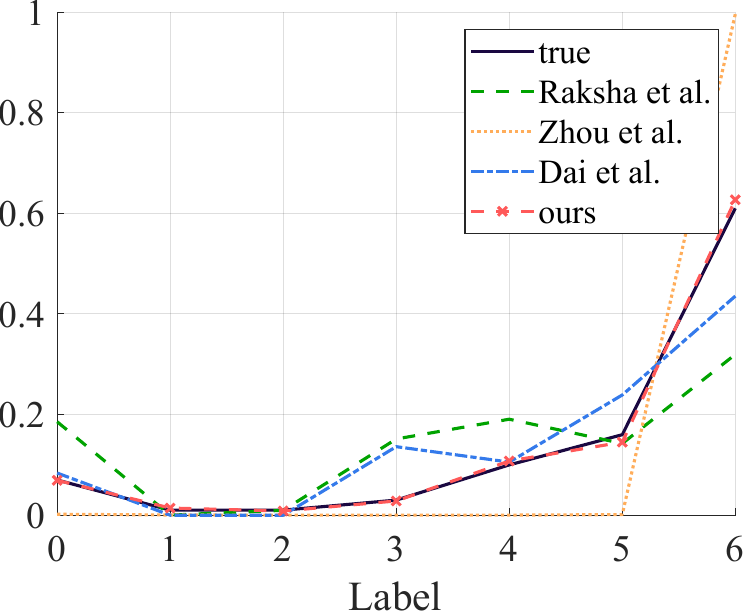}
        \caption*{$\alpha=2$}
        \label{fer2}
    \end{minipage}
    \hfill
    \begin{minipage}[b]{0.24\linewidth}
        \includegraphics[width=\linewidth]{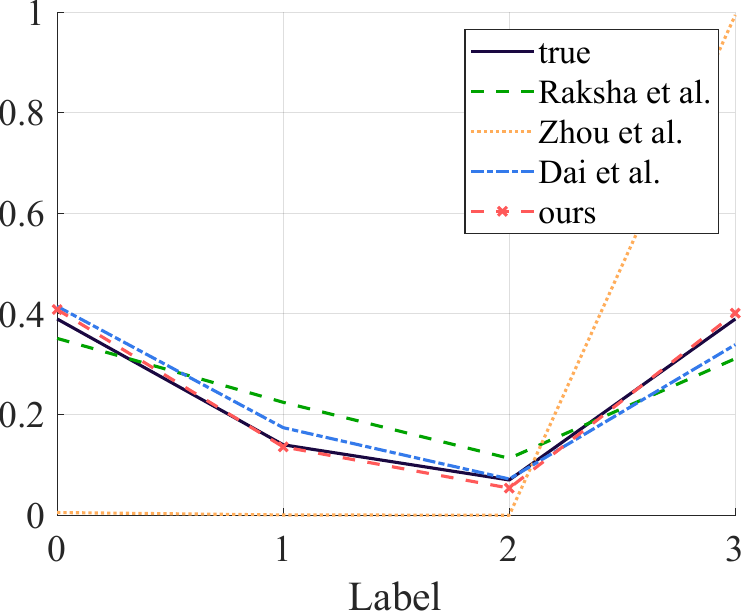}
        \caption*{$\alpha=1$}
        \label{agnews1}
    \end{minipage}

    \captionsetup{justification=centering}
    \caption*{\hspace{3em}MNIST \hspace{5em} Fashion-MNIST \hspace{4em} Fer2013 \hspace{5em} AG-News}

    \caption{Predicted distribution in non-IID of distribution-based label imbalance under different \textbf{$\alpha$}.}

    \label{noniid_D}
    
\end{figure*}

The Figure \ref{noniid_D} offer a comprehensive view of the performance under varying degrees of data heterogeneity induced by changing the Dirichlet parameter alpha $\alpha$. 
Notably, our method consistently demonstrates a superior ability to adhere closely to the true distributions across all datasets and $\alpha$ settings. This indicates that our model maintaining high performance even as data heterogeneity increases.
The reveals that all methods struggle to some extent as heterogeneity intensifies. However, our method exhibits a robust response, maintaining closer alignment with the true distribution compared to SOTA. This suggests that our model is particularly well-suited for federated learning applications where client data may vary dramatically, ensuring consistent performance across a range of complex scenarios.
Even in datasets with fewer labels like Fer2013 and AG-News, where modeling might theoretically be simpler, our method still proves effective, handling nuances in data distribution with finesse.

\subsection{Inference under Defenses}

Local differential privacy (LDP) offers robust mathematical guarantees for data privacy in federated learning by introducing random noise into the results of data publication or queries, thereby protecting individual information from inference. 
% The application of differential privacy in federated learning is generally divided into Central Differential Privacy (CDP) and Local Differential Privacy (LDP). 
% CDP is typically implemented on the server side, where noise is added after aggregating updates (such as model weights or gradients) uploaded from multiple clients. This method protects the privacy of the entire dataset, ensuring that no individual information can be accurately identified from the aggregated data. In contrast, 
% LDP is implemented on the client side, with each participant adding noise to their data before transmission. Even if the data collected by the server is leaked, it remains difficult to trace back to any individual, thus providing stricter privacy protection.
% Therefore, in this chapter, 
We validate the impact of local differential privacy techniques on the inferential capabilities of the proposed model. We focus on assessing the variations in model performance under different privacy budgets and the specific impact of privacy protection measures on model efficacy. Through these experiments, we aim to demonstrate that even under stringent privacy constraints, the label distribution inference model can still achieve satisfactory performance standards.

\begin{table}[h]
\centering
\caption{Performance Metrics at Various Epsilon Values.}
\label{dd}
\scalebox{1}{
\begin{tabular}{c|c|c|c|c|c}
\hline
$\epsilon$ & 1 & 2 & 5 & 10 & 40 \\ \hline
Wass    & 0.3738     & 0.2248     & 0.2041     & 0.2055      & 0.1451      \\ 
KL     & 0.0689     & 0.0382     & 0.0256      & 0.0257      & 0.0208      \\ 
JS      & 0.0194     & 0.0112     & 0.0071     & 0.0070      & 0.0058      \\ 
L1      & 0.2606     & 0.1646     & 0.1483     & 0.1425      & 0.1252      \\ 
ACC($\%$) & 50.46      & 77.25      & 85.66      & 87.78       & 88.55       \\ \hline
\end{tabular}}
\end{table}

\begin{figure*}[ht]
    \centering
    \begin{minipage}[b]{0.24\linewidth} % Adjusted width of each minipage to fill the double column
        \includegraphics[width=\linewidth,height=0.87\linewidth]{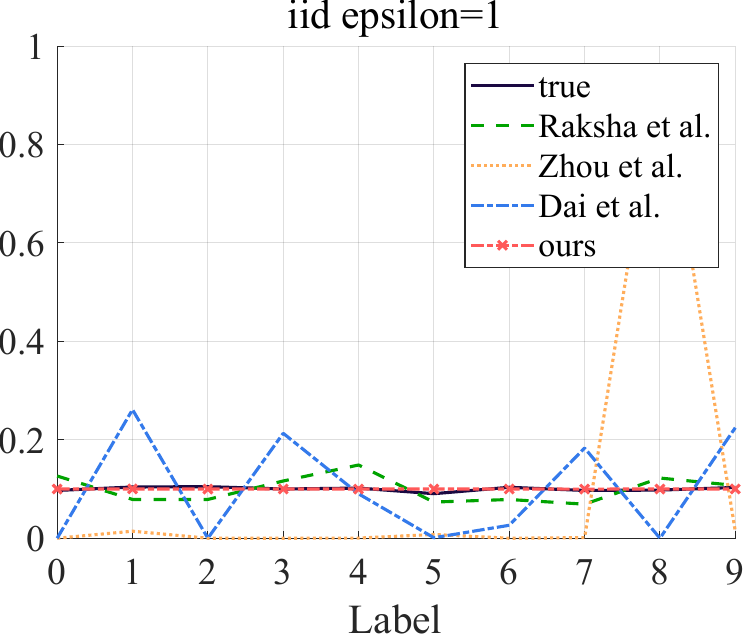}
        % \subcaption{IID}
        \label{1a}
    \end{minipage}
    \hfill
    \begin{minipage}[b]{0.24\linewidth}
        \includegraphics[width=\linewidth, height=0.87\linewidth]{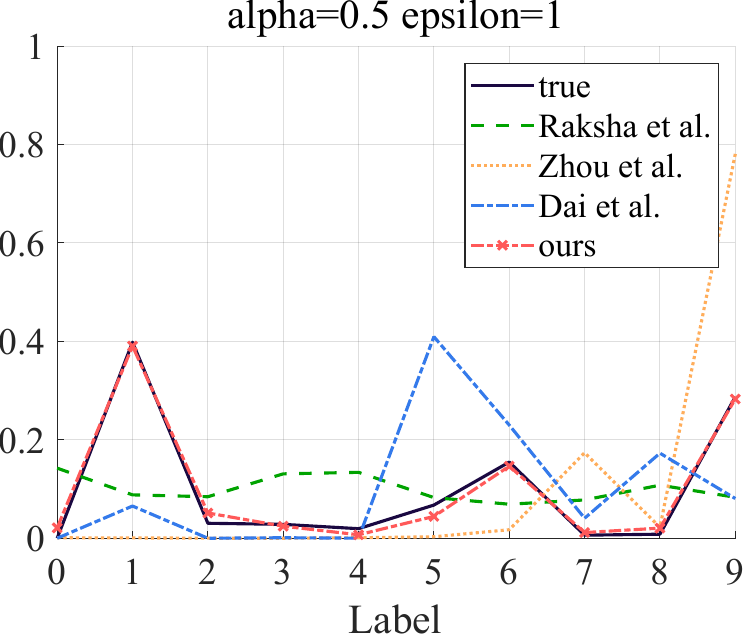}
        % \subcaption{Non-IID with $alpha=0.5$}
        \label{1b}
    \end{minipage}
    \hfill
    \begin{minipage}[b]{0.24\linewidth}
        \includegraphics[width=\linewidth, height=0.87\linewidth]{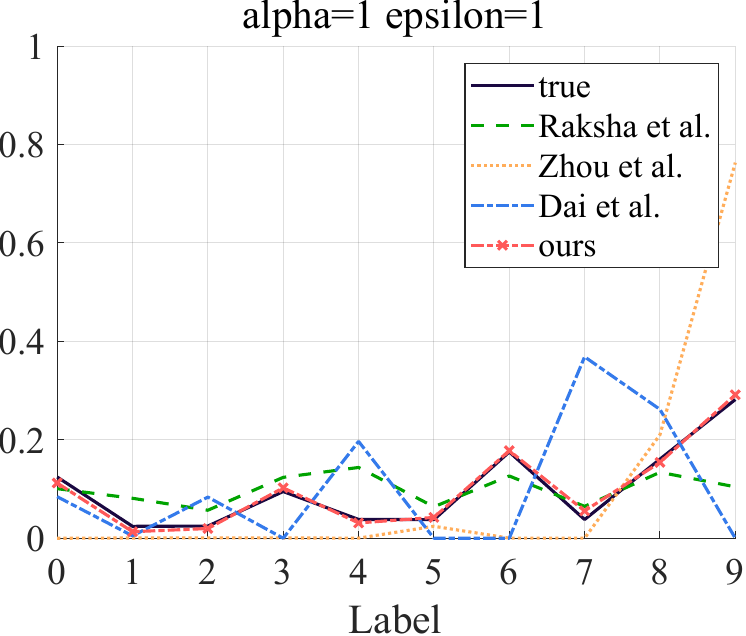}
        % \subcaption{Non-IID with $alpha=1$}
        \label{1c}
    \end{minipage}
    \hfill
    \begin{minipage}[b]{0.24\linewidth}
        \includegraphics[width=\linewidth, height=0.87\linewidth]{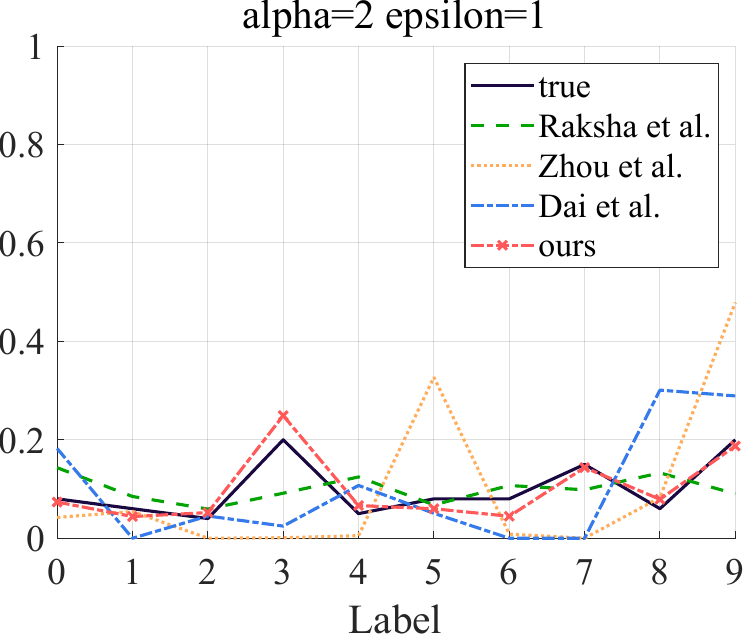}
        % \subcaption{Non-IID with $alpha=2$}
        \label{1d} % Corrected label for unique referencing
    \end{minipage}
    \caption{Predicted distribution attack against differential privacy under Non-IID.}
    \label{dp}
\end{figure*}

Table \ref{dd} illustrates the impact of our proposed method under five different $\epsilon$ settings for differential privacy defense. As the privacy budget $\epsilon$ decreases, the distance between the predicted distribution and the actual label distribution of the target client relatively increases, but this corresponds to a significant reduction in the model's usability. When the privacy budget is $5$, the L1 distance is $0.1476$, and as the privacy budget reduces to $1$, the L1 distance increases to $0.2455$. However, the victim client model's prediction accuracy for the main task drastically decreases from $85.66\%$ to $50.46\%$. Considering the impact of differential privacy on model usability, we believe that the fluctuations in the precision of label distribution predictions by the attack model are acceptable. Additionally, the server's ability to obtain the epsilon value for local differential privacy and to add corresponding noise in virtual clients can mitigate the effects of differential privacy defense strategies on the inference model.

Figure \ref{dp} presents the results of our method compared to SOTA methods under varying levels of data heterogeneity, with a privacy budget of $\epsilon=1$. The results demonstrate that, both in IID and Non-IID scenarios, our method more precisely predicts the actual label distributions than the SOTAs.
The enhanced accuracy is primarily attributed to our approach of simulating multiple virtual clients with different data distributions, whose temporal generalization performance data is used to train the inference model. In contrast, the methods by Dai et al. and Raksha et al. primarily infer by analyzing changes in the model's output layer parameters, that is more susceptible to noise introduced by differential privacy strategies. The presence of differential privacy noise leads to significant deviations in their inference results from the actual label distributions, especially in environments with high data heterogeneity.
Our research indicates that analyzing the generalization performance of virtual clients can effectively mitigate the adverse impacts of differential privacy noise on inference accuracy. This capability allows for maintaining precise inference models while upholding high levels of privacy protection. This finding is crucial for designing efficient and privacy-preserving data analysis strategies in practical federated learning applications.

% \subsection{Impact of dataset size}
% \begin{table}[h]
%     \centering
%     \caption{Impact of number for virtual clients.}
%     \label{datasize}
%     \scalebox{0.85}{
%     \begin{tabular}{c|c|c|c|c}
%         \hline
%         Virtual clients & Wass  & KL    & JS    & L1    \\
%         \hline
%         100     & 0.2693 & 0.0407 & 0.0112 & 0.1748 \\
%         300     & 0.1962 & 0.0216 & 0.0062 & 0.1258 \\
%         500     & 0.1276 & 0.0141 & 0.0039 & 0.1056 \\
%         % 800     & 0.1564 & 0.0135 & 0.0953 & 0.0037 \\
%         1000    & 0.1275 & 0.0106 & 0.0030 & 0.0840 \\
%         1200    & 0.1087 & 0.0097 & 0.0027 & 0.0772 \\
%         1500    & 0.1036 & 0.0070 & 0.0019 & 0.0663 \\
%         \hline
%     \end{tabular}}    
% \end{table}

% We increased the number of virtual clients constructed under each data distribution scenario from 100 to 1500 to evaluate the impact of dataset size on the prediction of the target client's label distribution. As shown in Table \ref{datasize}, the distance between the predicted label distribution and the actual label distribution of the target client gradually decreases as the number of virtual clients increases. This result indicates a positive correlation between the number of virtual clients and the effectiveness of the inference model. However, since a greater number of virtual clients demands more computational resources and time, practical deployment may require selecting an appropriate dataset size to balance the trade-off between inference accuracy and computational cost.

\section{Conclusion}
\label{5}
This paper introduces a novel label distribution inference attack method for federated learning environments, which leverages temporal variations in the model's generalization performance across different labels. By simulating virtual clients and analyzing the model parameters uploaded by real clients, this method infers the label distribution of their data. Tested across multiple datasets, this study not only confirms the efficacy of the method but also demonstrates its robustness against differential privacy protection measures.

This advancement holds significant value for practical applications, especially in industries where data sensitivity is paramount, such as healthcare and financial services. For instance, in healthcare, accurately understanding the distribution of diseases among populations can optimize resource allocation and predict epidemic trends, but it may also pose risks to patient data privacy. However, several questions remain for future exploration, such as the feasibility of deploying attackers on clients and extracting information from the global model through active attacks. We leave these issues to future research.

\printbibliography

\end{document}